\documentclass{article}

\usepackage[final]{neurips_2024}
\usepackage[utf8]{inputenc} 
\usepackage[T1]{fontenc} 
\usepackage{hyperref}   
\usepackage{url}       
\usepackage{booktabs} 
\usepackage{amsfonts}
\usepackage{nicefrac} 
\usepackage{microtype} 
\usepackage{xcolor}   
\usepackage{graphicx}
\usepackage{multirow}
\usepackage{multicol}
\usepackage{subcaption}
\usepackage[inline]{enumitem}
\definecolor{lowdrop}{HTML}{B7E1CD}

\title{Communication Compression for Tensor Parallel LLM Inference}

\author{%
Jan Hansen-Palmus$^{1,2\thanks{Research conducted during internship at Recogni}}$ \quad Michael Truong Le$^1$ \quad Oliver Hausdörfer$^2$ \quad Alok Verma$^1$\\
$^1$Recogni \quad $^2$Technical University of Munich\\
\texttt{\{jan.hansenpalmus,michael,alokge\}@recogni.com}\\
\texttt{oliver.hausdoerfer@tum.de}
}

\begin{document}
\maketitle

\begin{abstract}
Large Language Models (LLMs) have pushed the frontier of artificial intelligence but are comprised of hundreds of billions of parameters and operations. For faster inference latency, LLMs are deployed on multiple hardware accelerators through various Model Parallelism strategies. Our paper looks into the details on one such strategy - Tensor Parallel - and proposes to reduce latency by compressing inter-accelerator communication. We leverage fine grained quantization techniques to compress selected activations by 3.5 - 4.5x. Our proposed method leads up to 2x reduction of time-to-first-token (TTFT) with negligible model performance degradation. 
\end{abstract}

\section{Introduction}
Large Language Models (LLMs) have become essential across various applications due to their exceptional performance. As model performance tends to improve with increased parameter counts, LLMs have been significantly scaled in recent years, with contemporary models now reaching 500B+ parameters \citep{palm}.

Deploying such large models for inference presents major challenges \citep{efficientlyscaling}. Tensor Parallel \citep{megatron} addresses this by splitting layers on multiple accelerators, enabling the execution of extremely large models and significantly reducing latency. However, Tensor Parallel demands accumulation of results from accelerators, as shown in Figure \ref{fig:highlight_flow}, and can lead to data communication bottlenecks \citep{optimizingcommunication, mnemosyne}, especially during the first auto-regressive inference step (the prefill phase).

One approach to mitigate these bottlenecks, and thus reduce model latency even further, is to quantize activations before communication, which reduces the time needed to accumulate results from accelerators in a Tensor Parallel group. However, the presence of outliers \citep{llmint8, awq} complicates this strategy, necessitating fine-grained quantization approaches. We leverage such approaches proposed by \cite{microscaling} to compress activations and demonstrate the potency of communication compression by measuring time-to-first-token (TTFT) in realistic inference scenarios using different inference hardware setups. We find that for hardware setups which have slower inter-accelerator bandwidths, the TTFT can be improved by 3.5 - 4.5x with negligible degradation of model performance.

\section{Background}
\subsection{Parallel Inference}

Given the massive parameter and operations requirement of LLMs \citep{llama3, llmsurvey}, distributing their parameters and operations across a large number of accelerators is essential, which is achieved through different parallelism techniques. Among common parallelism techniques used to fit LLMs on multiple accelerators for inference \citep{megatron, reducingactivationrecomputation, zero, optimizingcommunication}, Tensor Parallel (TP) is usually the most widely used one because it allows a very effective way of reducing latency and scaling down model size per accelerator. Other parallelism strategies, such as Pipeline Parallelism or Sequence Parallel, are combined with TP to improve further upon it. TP has two possible sub-variants: Column-wise and Row-wise parallelism \citep{megatron}, enable the partitioning of linear layers along the column or row dimension of weights respectively.

While Tensor Parallel offers significant benefits, it also increases communication overhead, as results computed on individual accelerators must be frequently synchronized using collective operations \citep{mpimessagepassinginterfacestandard}, \citep{nccl}. This becomes particularly problematic during the prefill phase, where activation tensors for the entire token sequence need to be transmitted. This can lead to communication bottlenecks \citep{optimizingcommunication, mnemosyne}, especially if the inter-accelerator bandwidth is low. While \cite{tpcompression} explored various approaches to reduce communication overhead during training, to the best of our knowledge, this issue has not yet been addressed at inference-time by the research community.

\subsection{Model Quantization}
The quantization of weights and/or activations of LLMs \citep{atom, awq,flexgen,gear, smoothquant, rptq, gptq} has been a very active field of research. Quantization of weights to lower bit-widths reduces the memory needed to store parameters and also improves throughput due to better memory bandwidth utilization. Furthermore, quantization of activations and also computations allows reducing activation memory footprint and improves throughput as well by utilizing faster and cheaper computation engines on recent accelerators \citep{blackwell}. 

Unlike weights, activations are hard to quantize due to their dynamic nature \citep{atom, flexgen} and the presence of outliers, which has been extensively studied in literature \citep{llmint8, awq, smoothquant}. \cite{llmint8} found that accurately representing these outliers is critical for maintaining model performance: Even though outliers represent only a small fraction of input features, removing them leads to a significant degradation in Perplexity.

A common approach to activation quantization is to normalize the values based on the tensor’s absolute maximum. However, this can be problematic due to the presence of outlier values. Using the absolute maximum, which is dominated by these outliers, leads to poor representation of the remaining values in the tensor, as they become compressed into a narrow range and lose precision.

To address this issue, various solutions have been developed. One approach is to limit the impact of outliers by grouping tensor values and quantizing them together, which helps reducing the quantization error. Mixed precision methods 
 \citep{atom, llmint8, gear, kvquant}, on the other hand, distinguish between outliers and non-outliers, using different data types for each group during quantization. 

Since we compress activations of specific layers, in the following section, we briefly review methods related to low-bit activation quantization.

\section{Related Work}
\subsection{KV-cache Quantization}
LLM inference involves generating tokens in an auto-regressive manner, and the majority of the computation time is spent accessing the KV-cache from memory \citep{kvquant, kivi}. KV-cache quantization addresses this issue by quantizing Key and Value activations, reducing the effective size of the KV-cache.

\cite{kivi} observed distinct outlier characteristics between Key and Value caches and apply per-channel and per-token quantization respectively. Per-channel quantization groups and quantizes values within the same channel while per-token quantization applies this to token dimension. \cite{flexgen} employs group-wise quantization, which takes per-channel quantization a step further by grouping small blocks of values within each channel and quantizing them independently, which reduces quantization errors. \cite{kvquant, gear} propose extracting a sparse outlier matrix composed of the top 1-2\% of the largest values, which are preserved in higher precision.

\begin{figure}
\centering
    \begin{subfigure}{.51\textwidth}
        \centering
        \includegraphics[width=1\linewidth]{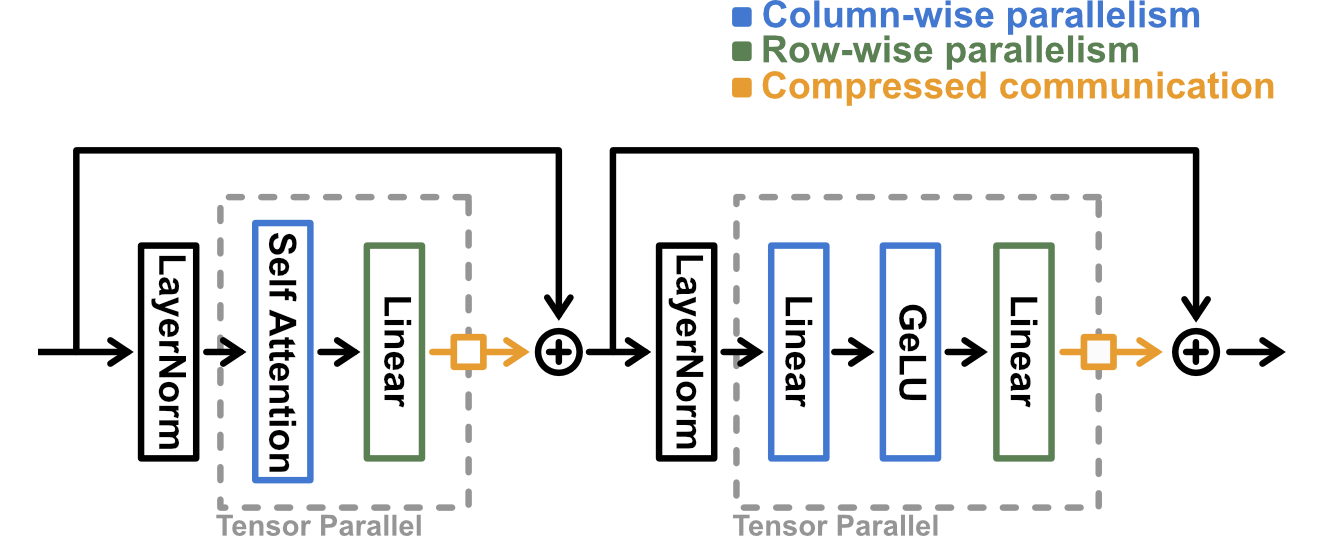}
        \caption{}
        \label{fig:model_flow}
    \end{subfigure}\hfill
    \begin{subfigure}{.48\textwidth}
        \centering
 \includegraphics[width=1\linewidth]{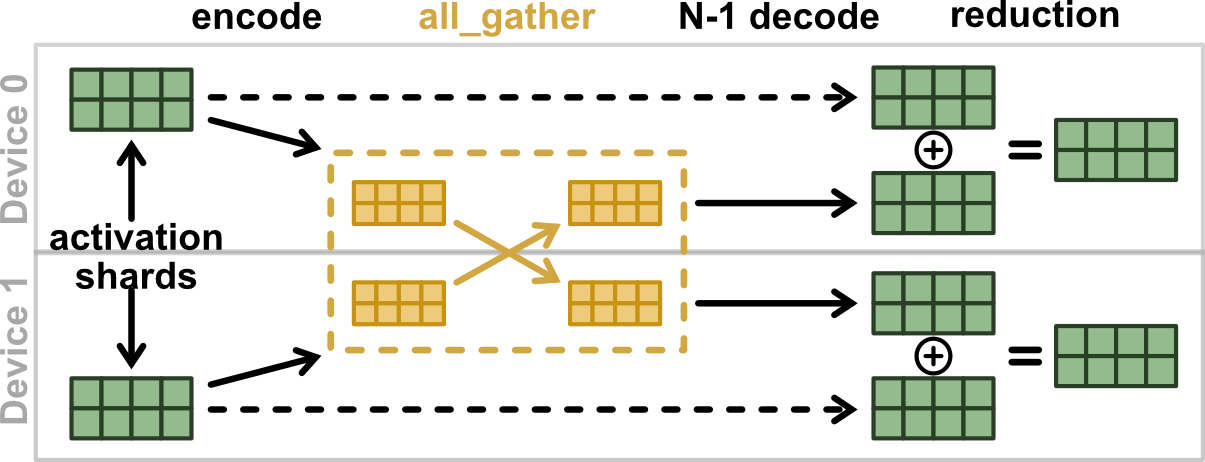}
        \caption{}
        \label{fig:flow}
    \end{subfigure}
\caption{An illustration of transformer-based LLM model parallelized using TP. In Figure \ref{fig:model_flow}, column-wise and row-wise TP layers are marked blue and red, respectively. Before reduction, we propose to compress the all\_gather collective op (orange in Figure \ref{fig:model_flow}), as presented in Figure \ref{fig:flow}.}
\label{fig:highlight_flow}
\end{figure}

Unlike KV-cache activation compression, we target communication compression occurring after execution of row-wise TP linear layers, as shown in Figure \ref{fig:model_flow} and \ref{fig:flow}. Compression and decompression of communication has to be done at much lower latency, otherwise improvements achieved by communication compression are offset by encoding and decoding steps.

\subsection{Weight and Activation Quantization}
The primary objective of weights and activation quantization is to reduce the overall model size while accelerating inference \citep{understanding, atom} by the use of specialized hardware, such as Nvidia's INT4 Tensor Cores \citep{blackwell}.

\cite{atom} proposed a mixed-precision technique that addresses the challenge of activation quantization by representing the 128 largest outlier channels with 8 bits per value. \cite{ocp_specification} introduces low-bit data formats applicable to matrix multiplication operations. In this method, a block of values is encoded in a low-bit floating-point format along with a shared exponent. Given the strong performance of \cite{ocp_specification} \citep{microscaling} data types, we based our work on their provided code \citep{mxgithub} and apply low-bit data types for activation compression experiments as well.

\section{Method}
\subsection{Communication Compression}\label{sec:method_communication_compression}
Figure \ref{fig:model_flow} highlights our approach to communication compression in a TP setting. We aim to compress the partial results of each worker after row-wise TP linear layers, and decompress them before reduction in each worker. Since our method introduces extra computations which offsets slower inter-accelerator communication, we must strike a balance between computation and compression. We found that block-wise quantization as proposed in \cite{ocp_specification} for low-bit compression is a strong contender to balance quantization error and compression latency. So, we evaluate a variety of low-bit data types using this fine-grained block-wise quantization scheme.

We also extended the data types available in \citep{mxgithub} to experiment with more variety of bit-widths to test higher compression. The following data types and parameters were evaluated:
\begin{enumerate}
\item \textbf{Value data types}: FP5 (E3M1, E2M2, E1M3), FP4 (E2M1, E1M2), FP2 (E1M1), INT3, INT4, INT5
\item \textbf{Block size}: 8, 16, 32
\item \textbf{Scale data type}: E8M0, E7M0, E6M0, E5M0, E4M0
\end{enumerate}

\subsection{Model Evaluation}
We measure the performance of compression approaches based on the compression rate, which is measured by the number of \textit{effective bits} \citep{gptq, awq}, and the increase in Perplexity metric \citep{kvquant, atom, llmint8, outliersuppression}, relative to the compression-free model with 16 bit (FP16) activations. We show that each possible hyper-parameter of quantization data type proposed in \cite{ocp_specification} can have varied effect on latency and model's Perplexity in Section \ref{sec:ablation}.

\subsection{Profiling}
To show we can reduce communication bottlenecks by utilizing our most performant quantization approaches, we measure the TTFT of models of different sizes in a deployment scenario. We base our profiling on the code provided by \cite{ibmfms} which uses \verb|torch.compile| \citep{torchcompile} to speed up inference of Llama 2 models along with TP. The architecture of the Llama 2 model family is very similar to other state-of-the-art LLMs and therefore provides valuable compression insights. We extended the code to add communication compression. In our profiling setup, each worker in a TP group of $N$ workers compresses output activations of each row-wise linear layer before communication, and decompresses N-1 activations gathered from all other workers. We finally reduce the decompressed activations using \verb|torch.sum| as shown in Figure \ref{fig:flow}.

\section{Experiments}

\subsection{Optimal Compression Scheme Search}\label{sec:ablation}
To pick an optimal compression scheme for each model we first evaluate the Perplexity of various state-of-the-art LLMs on 10\% of the Wikitext train dataset, which we deemed sufficient for this purpose, \cite{wikitext} by testing a subset of combinations of data types and block sizes. Please refer to \ref{appendix:ablation} for further details how we restricted the search space of quantization hyper-parameters prior to these experiments.

Next, we pick the final compression schemes for each model using the results from Table \ref{table:compressed_models_perplexity}: To ensure minimal model performance degradation, we consider only combinations which lead to an increase in Perplexity of less than 3\%. From the remaining candidates, we then choose the one with the largest compression ratio (lowest effective bits). This procedure ensures a good balance between compression ratio and quantization accuracy. Finally, to validate the performance of our chosen compression schemes, we evaluate them on the entire Wikitext test set (Table \ref{table:final_evaluation}). We observe minimal performance degradation with respect to 16 bit uncompressed communication, while being able to compress tensors by a factor of roughly $3.3$X.

\begin{table}[h]
\centering
\begin{tabular}{l c c c c c c c c c c}
\toprule
\multirow{2}{*}{\textbf{Value}} & \multirow{2}{*}{\textbf{Block}} & \multirow{2}{*}{\textbf{Eff.}} & \multicolumn{2}{c}{\textbf{Llama 3.1}} & \multicolumn{2}{c}{\textbf{Gemma2}} & \multicolumn{3}{c}{\textbf{Mistral}} \\ \cmidrule(lr){4-5} \cmidrule(lr){6-7} \cmidrule(lr){8-10}
                                 \textbf{Dtype}     &                           \textbf{Size}           &                                          \textbf{Bits} & \textbf{8B} & \textbf{70B} & \textbf{2B} & \textbf{9B} & \textbf{7B} & \textbf{22B} & \textbf{123B} \\ \midrule
FP16 & - & 16 & 7.12 & 4.24 & 14.50 & 10.17 & 5.55 & 4.16 & 2.66 \\ \midrule       
\multirow{3}{*}{FP3}             & 8                                    & 3.6                                       & 9.39\%       & 11.42\%      & 12.17\%      & 5.49\%        & 3.40\%            & 9.24\%          & 9.29\%       \\ 
                                      & 16                                   & 3.3                                       & 13.87\%       & 14.78\%       & 19.59\%       & 7.10\%        & 4.46\%            & 11.79\%         & 11.79\%          \\ 
                                      & 32                                   & 3.2                                       & 19.67\%       & 18.50\%       & 31.07\%       & 13.51\%       & 5.50\%            & 13.23\%         & 14.58\%          \\ \midrule
\multirow{3}{*}{FP4}             & 8                                    & 4.6                                       & 2.92\%  & 3.91\%        & 3.91\%        & 1.47\%  & 1.27\%  &  2.36\%  & 3.39\%           \\ 
                                      & 16                                   & 4.3                                       & 3.01\%        & 3.85\%        & 4.64\%        &  1.94\%  & 1.31\%  & 6.58\%          & 3.41\%           \\ 
                                      & 32                                   & 4.2                                       & 3.37\%        & 4.14\%        & 6.30\%        &  2.20\%  &  1.22\%  & 6.69\%          & 3.53\%           \\ \midrule
\multirow{3}{*}{FP5}             & 8                                    & 5.6                                       &  0.58\%   & 1.09\%        &  0.85\%  & -0.19\% &  0.49\%  & 0.30\%  & 0.69\%           \\ 
                                      & 16                                   & 5.3                                       & 0.57\%   & 1.14\%        &1.17\%  & 0.56\%  & 0.52\%          & 0.35\%          & 0.66\%           \\ 
                                      & 32                                   & 5.2                                       & 0.70\%   &  1.22\%   &  1.54\%  & 1.02\%  & 0.49\%          &  0.44\%  & 0.59\%           \\ \bottomrule
\end{tabular}
\caption{Perplexity evaluation is done on 10\% of the Wikitext2 training set and the degradation is reported relative (in \%) to the absolute FP16 performance in the first row. Grid search over block size and effective bits are done which show a significant effect on model performance. For FP3, FP4 and FP5, we chose E1M1, E2M1 and E2M2 sub-variants respectively.}
\label{table:compressed_models_perplexity}
\end{table}

\begin{table}[h]
\centering
\begin{tabular}{l l c c c c c}
\toprule
&&&&& \multicolumn{2}{c}{\textbf{Perplexity}} \\ 
\textbf{Model} & \textbf{Sub-variant} & \textbf{Value Dtype} & \textbf{Block Size} & \textbf{Bits} & \textbf{FP16} & \textbf{Increase}  \\ 
\midrule
\multirow{2}{*}{Llama 3.1} & 8B  & FP4  & 8 & 4.6 & 7.22 &3.22\%  \\ \cmidrule(lr){2-7}
                                    & 70B  & FP5  & 32 & 5.2 & 3.86 &1.68\%  \\ 
\midrule
\multirow{2}{*}{Gemma 2}   & 2B  & FP5  & 32  & 5.2& 14.27 &1.39\%  \\ \cmidrule(lr){2-7}
                                    & 9B  & FP4  & 32 & 4.2 & 10.40 &1.83\%  \\ 
\midrule
\multirow{4}{*}{Mistral}   & 7B  & FP4  & 32  & 4.2& 5.23 &1.18\%  \\ \cmidrule(lr){2-7}
                                   & 22B & FP4  & 8  & 4.6 & 4.02 &1.62\%  \\
                                     \cmidrule(lr){2-7}
                                    & 123B & FP5 & 32 & 5.2 & 2.65 &0.48\%  \\ 
\bottomrule
\end{tabular}
\caption{Evaluation of the best-performing quantization schemes on Wikitext2 test set. The Perplexity column shows the relative degradation of the compressed model performance compared to uncompressed FP16 model. As in Table \ref{table:compressed_models_perplexity}, for FP4 and FP5 the sub-variants E2M1 and E2M2 are chosen, respectively.}
\label{table:final_evaluation}
\end{table}

\subsection{Measuring TTFT Speedups}

\begin{table}[h]
\centering
\begin{tabular}{llcccc}
\toprule
                &                   &                   & \multicolumn{2}{c}{\textbf{TTFT [s]}}                         &  \\ 
\textbf{Model}  & \textbf{Accelerators} & \textbf{Input}    & \textbf{Uncompressed} & \textbf{Compressed}    & \textbf{Speedup} \\ 
\midrule
\multirow{4}{*}{LLama 2 70b}  & \multirow{2}{*}{8xL4} & 2x64  & 0.58 (0.04) & 0.32 (0.68) & 1.83\\ 
                            &                                   & 2x128 & 1.07 (0.04) & 0.52 (0.63) & 2.08 \\ \cmidrule{2-6}
                            & \multirow{2}{*}{4xA100} & 2x128  & 0.09 (0.00) & 0.15 (0.00) & 0.56 \\ 
                            &                                   & 2x256 & 0.13 (0.00) & 0.19 (0.00) & 0.70 \\ \midrule
\multirow{2}{*}{Llama 2 13b}  & \multirow{2}{*}{4xL4} & 8x128 & 0.67 (0.00) & 0.33 (0.00) & 2.05 \\ 
                            &                                   & 8x256 & 1.37 (0.00) & 0.70 (0.00) & 1.96 \\ 
                            \midrule
\multirow{2}{*}{Llama 2 7b}   & \multirow{2}{*}{2xL4} & 16x128 & 0.39 (0.00) & 0.45 (0.00) & 0.88 \\ 
                            &                                   & 16x256 & 0.79 (0.00) & 0.77 (0.00) & 1.03 \\ 
\bottomrule
\end{tabular}
\caption{Inference profiling results are shown for different models and TP configurations. For compression, quantization scheme with value data type FP4 E2M1, batch size 32 and scale data type E8M0 was used, which has 4.25 effective bits. The TTFT values correspond to the median of 32 model forward passes, additionally we report the standard deviation in brackets.}
\label{table:profiling}
\end{table}

We demonstrate the potential speedups achievable by compressing communication in \ref{table:profiling}.
As \cite{ibmfms} only supports the Llama 2 model family, we decided to conduct measurements using the FP4 quantization scheme deemed best by \cite{microscaling}, which has a similar compression ratio to the schemes we picked in \ref{sec:ablation}. We achieve a speed-up between 2x - 1.2x based on the TP degree and hardware setup. We profile our method on cutting-edge NVIDIA accelerators: L4, and A100 which are accessed through Google Cloud Platform \citep{gcp}. L4 GPUs in a node are connected with PCIe Gen4 x16 and have 64GB/s bandwidth, while A100 have 600 GB/s bidirectional any-to-any bandwidth. These both types of GPU provide good coverage of computing throughput(FLOPs/sec) and GPU-GPU bandwidth to showcase communication compression benefits.
Generally, setups containing more than two L4 GPUs greatly benefit from our approach of communication compression. Due to the rather slow interconnect, these setups are heavily bottle-necked by communication overhead. In contrast, our A100 setup is slowed down by our quantization scheme: Due to the fast interconnect, the quantization overhead is so large that the we don't benefit from data compression.

\begin{table}[h]
\centering
\begin{tabular}{lccccc}
\hline
\textbf{} & \multicolumn{3}{c}{\textbf{Perplexity}} & \multicolumn{2}{c}{\textbf{TTFT} (Llama 2 70B)} \\ 
\textbf{} & Llama 3.1 8B & Gemma2 2B & Mistral 7B & 8xL4& 4xA100\\ \hline
FP16 & 7.22 &  14.27 & 5.23 & 1.06s & 0.13s\\ \hline
MX4 E2M1 & 3.22\% & 6.06\% & 1.17\% & 2.07x &  0.70x \\ \hline
Int4 & 6.19\% & 8.83\% & 15.05\% & 2.60x & 0.95x \\ \hline
TopK 3x & 115.54\% & 80.17\% & 21.38\% & 1.80x & 0.55x \\ \hline
\end{tabular}
\caption{Comparison with methods from \cite{tpcompression}. MX4 E2M1 (batch size 32 and E8M0 as scaling dtype) is compared to channel-wise INT4 and TopK compression using the Wikitext2 test set. The speedup of the compressed model is compared to the uncompressed FP16 model, we present absolute numbers in the first row. Input shapes 2x128 and 2x256 were used for the setups 8xL4 and 4xA100 respectively.}
\label{table:sota_comparison}
\end{table}
\subsection{SoTA Comparison}

We compare our compression method with the relevant state-of-the-art method described in \citep{tpcompression} which proposes multiple learned and non-learned methods for communication compression. Since our method targets inference-only non-learned optimization, we compare to the two fastest non-learned approaches: channel-wise INT quantization, and TopK compression which keeps the K largest magnitudes and zeroes all other values.

In Table \ref{table:sota_comparison}, we show that both of their compression techniques are lead to much higher degradation of Perplexity metric. INT4 compression offers substantial speedups due to the minimal computational overhead, but shows higher Perplexity degradation when compared to fine-grained quantization approach. Improving the accuracy of this approach is left for future work.

\section{Conclusion and Limitations}
Our work builds on the foundation set by \citet{microscaling} and \citet{ocp_specification} by incorporating additional data types and tuning parameters to optimize compression. To achieve this, we developed a model-dependent hyper-parameter selection procedure that effectively balances the compression rate with model accuracy. The proposed compression method reduces activation sizes by a factor of 3.5 to 4.5x, with minimal degradation in performance which results in an improvement of LLMs’ TTFT by a factor of 1.2 to 2x, depending on the hardware setup.

While our compression methods shows substantial TTFT improvements for some hardware configurations, it does not lead to improvement when inter-accelerator bandwidth is substantially high and inference is not bandwidth-bound anymore. Furthermore, the latency introduced by compression algorithms can offset any communication compression speedup, so compressing more but with slower algorithms is not helpful. It could be possible to accelerate compression by utilizing specialized hardware, but such strategies are beyond the current scope and are left for future 
investigation. Furthermore, by improving model quantization techniques and adopting 8 bit matrix multiplications, the communication size of TP linear layers could be reduced easily without extra compression costs.  

Our current profiling setup could also be enhanced further to take into account in-flight batching and other parallelism strategies and provide more extensive details of throughput improvements.

\begin{ack}
We acknowledge the insights offered by Frederik Gerzer and Thomas Pfeil, which helped shape the experiments presented in this paper. We are also grateful to the engineering and research teams at Recogni who provided necessary infrastructure to run various experiments on cloud machines easily.
\end{ack}

\bibliography{citations}

\small

\appendix
\section{Appendix / supplemental material}
\subsection{Ablation over quantization hyper-parameters}\label{appendix:ablation}
We ablate over the extended space of possible data types and parameters (\ref{sec:method_communication_compression}). We aim at restricting our parameter space by picking parameters (Table \ref{table:mxdata_type_ablation}) which promise good quantization accuracy for different compression ratio requirements. As scale data type, E5M0 was generally the best option, as lower bits for scale introduce unacceptable Perplexity degradation while higher do not lead to any improvement. For value bits we found the data types FP4 E2M1 or FP5 E2M2 were sufficient to cover different required compression ratios, while FP3 E1M1 showed large Perplexity increase. Since the choice of the block size has a large effect on Gemma's performance, we decided to keep all possible parameters. We deem it necessary to evaluate all of these compression configuration because naively just choosing the highest compression configuration leads to high degradation of model Perplexity.

\begin{table}[h]
\centering
\begin{tabular}{ccccc}
\hline
\multirow{2}{*}{} & \multirow{2}{*}{\textbf{Parameter}} & \multicolumn{3}{c}{\textbf{Perplexity increase [\%]}} \\
                  &                                    & \textbf{Llama 3.1B} & \textbf{Mistral 7B} & \textbf{Gemma 2B} \\ \hline
\multirow{5}{*}{\textbf{Scale Bits}} 
                  & 4                                  & 3.67                & 6.11                & 7.46              \\ 
                  & 5                                  & 3.37                & 1.22                & 6.30              \\  
                  & 6                                  & 3.37                & 1.29                & 6.30              \\  
                  & 7                                  & 3.37                & 1.29                & 6.30              \\   \hline
\multirow{6}{*}{\textbf{Value Data Type}} 
                  & FP3 E1M1                           & 19.67               & 5.52                & 31.07             \\    
                  & FP4 E1M2                           & 4.45                & 1.27               & 9.95              \\   
                  & FP4 E2M1                           & 3.37                & 1.29                & 6.30              \\   
                  & FP5 E1M3                           & 0.90                & 0.38                & 3.35              \\  
                  & FP5 E2M2                           & 0.70                & 0.45                & 1.54              \\  
                  & FP5 E3M1                           & 2.63                & 1.18                & 3.02              \\  
                  & INT 3                              & 19.67               & 5.52                & 31.07             \\  
                  & INT 4                              & 4.45                & 1.27                & 9.95              \\  
                  & INT 5                              & 0.90                & 0.38                & 3.35              \\ \hline
\multirow{3}{*}{\textbf{Block Size}} 
                  & 8                                  & 2.92                & 1.27               & 3.91              \\ 
                  & 16                                 & 3.01                & 1.37                & 4.64              \\  
                  & 32                                 & 3.37                & 1.29                & 6.30              \\ \hline
\multirow{5}{*}{\textbf{Parallelism}} 
                  & 2                                  & 3.37                & 1.29                & 6.30              \\  
                  & 4                                  & 3.35                & 1.13                & 5.22              \\  
                  & 8                                  & 2.83                & 1.03                & 4.83              \\   
                  & 16                                 & 2.88                & 0.97                & 4.59              \\ 
                  & 32                                 & 2.88                & 0.95                & 3.80              \\ \hline
\end{tabular}
\caption{Ablation over scale bits, value bits and type, block size and degree of parallelism. Evaluation on 10\% of the Wikitext2 training set.}
\label{table:mxdata_type_ablation}  
\end{table}

\end{document}